\pdfoutput=1
\documentclass[twocolumn]{article}

\usepackage[T1]{fontenc}
\usepackage[lining]{XCharter}
\usepackage[xcharter,vvarbb]{newtxmath}
\usepackage[scale=0.92,lining,tabular]{roboto}
\usepackage[scale=0.86]{sourcecodepro}
\usepackage[protrusion=true,expansion=true]{microtype}
\usepackage[letterpaper,margin=0.75in,top=0.9in,bottom=0.9in,columnsep=0.3in,headsep=14pt,footskip=26pt]{geometry}
\usepackage{booktabs,array,amsmath,multirow,graphicx,subcaption}
\usepackage[dvipsnames]{xcolor}
\usepackage{enumitem,titlesec,fancyhdr,etoolbox}
\usepackage[most]{tcolorbox}
\usepackage[numbers,sort&compress]{natbib}

\definecolor{ink}{HTML}{23231E}
\definecolor{linkblue}{HTML}{283746}
\definecolor{muted}{HTML}{6E6E64}
\definecolor{hair}{HTML}{B4B4AA}

\usepackage[colorlinks=true,citecolor=linkblue,linkcolor=linkblue,urlcolor=linkblue,
  pdftitle={PhoneBench: Phone Agents Ace Existing Benchmarks but Fail Everyday Tasks},
  pdfauthor={Sabi AI Lab}]{hyperref}
\color{ink}

\titleformat{\section}{\sffamily\bfseries\large}{\thesection}{0.8em}{}
\titleformat{name=\section,numberless}{\sffamily\bfseries\large}{}{0pt}{}
\titlespacing*{\section}{0pt}{16pt plus 4pt minus 2pt}{6pt plus 2pt}
\titleformat{\subsection}{\sffamily\bfseries\normalsize}{\thesubsection}{0.8em}{}
\titlespacing*{\subsection}{0pt}{12pt plus 3pt minus 2pt}{4pt plus 1pt}
\titleformat{\paragraph}[runin]{\bfseries}{}{0pt}{}
\titlespacing*{\paragraph}{0pt}{6pt plus 2pt minus 1pt}{0.5em}

\usepackage[font=small,labelfont=bf,labelsep=period,skip=6pt,justification=justified,singlelinecheck=false]{caption}

\AtBeginEnvironment{tabular}{\let\tiny\countfont}
\AtBeginEnvironment{figure}{\let\fbox\softfbox}

\newif\ifnarrowfloat
\newcommand{\narrowfile}[1]{\AddToHook{file/#1/before}{\narrowfloattrue}\AddToHook{file/#1/after}{\narrowfloatfalse}}
\makeatletter
\def\pb@wideb{\@dblfloat{\pb@type}[tp]}\def\pb@widee{\end@dblfloat}
\def\pb@narrowb{\@float{\pb@type}[tbp]}\def\pb@narrowe{\end@float}
\renewenvironment{figure}[1][]{\def\pb@type{figure}\ifnarrowfloat\let\pb@e\pb@narrowe\expandafter\pb@narrowb
  \else\let\pb@e\pb@widee\expandafter\pb@wideb\fi}{\pb@e}
\renewenvironment{table}[1][]{\def\pb@type{table}\ifnarrowfloat\let\pb@e\pb@narrowe\expandafter\pb@narrowb
  \else\let\pb@e\pb@widee\expandafter\pb@wideb\fi}{\pb@e}
\makeatother
\setlist{itemsep=2pt,topsep=4pt,parsep=0pt}

\fancypagestyle{plain}{\fancyhf{}\fancyfoot[C]{\small\thepage}}

\renewenvironment{abstract}
  {\par\vspace{10pt}{\centering\bfseries Abstract\par}\vspace{4pt}
   \begin{list}{}{\setlength{\leftmargin}{0.45in}\setlength{\rightmargin}{0.45in}\setlength{\topsep}{0pt}}
   \item\relax\small}
  {\end{list}\par\vspace{8pt}}

\newcommand{\name}{PhoneBench}
\newcommand{\hard}{PhoneBench-Hard}

\definecolor{ink}{HTML}{111111}
\definecolor{linkblue}{HTML}{0068A8}
\definecolor{muted}{HTML}{555555}
\definecolor{hair}{HTML}{BBBBBB}
\color{ink}
\newcommand{\colref}{blue}\newcommand{\colours}{vermilion}\newcommand{\Colours}{Vermilion}
\makeatletter
\let\pb@includegraphics\includegraphics
\renewcommand{\includegraphics}[2][]{\IfFileExists{vivid/#2.pdf}{\pb@includegraphics[#1]{vivid/#2}}{\pb@includegraphics[#1]{#2}}}
\makeatother

\narrowfile{tables/benchmarks.tex}
\narrowfile{tables/audit.tex}
\narrowfile{tables/apps.tex}

\begin{document}
\twocolumn[{
\begingroup\raggedright
\raisebox{-1pt}{\includegraphics[height=12pt]{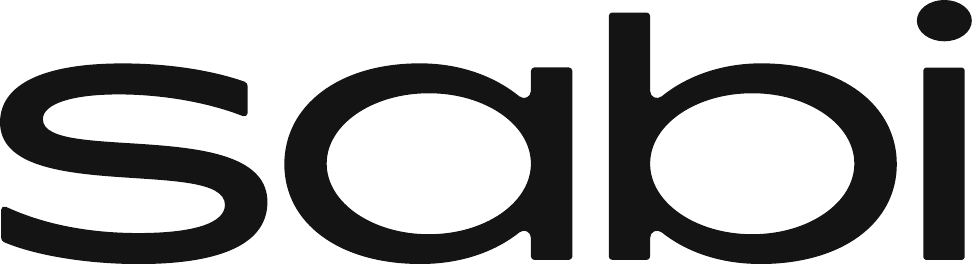}}\hfill{\small\textcolor{muted}{Technical report \,·\, November 2026}}\par
\vspace{22pt}
{\sffamily\bfseries\fontsize{21}{26}\selectfont \name: Phone Agents Ace Existing Benchmarks\\but Fail Everyday Tasks\par}
\vspace{12pt}
{\normalsize Owais Aijaz, Nikhil Ranjan, Ahmer Tabassum, Amish Gupta, Abhyuday Prasad, Siddhant Kharbanda,
Aryan Seth, Atmadeep Banerjee, Rahul Chhabra\par}
\vspace{3pt}
{\small\itshape Sabi AI Lab\par}
\endgroup
\begin{abstract}
Phone-use agents now reach 98\% on AndroidWorld and 82\% on MobileWorld's GUI tasks, yet the best model we test
solves 10.5\% of everyday multi-app tasks. We introduce \name, a benchmark of everyday phone requests on thirteen apps
in an Android emulator, including self-hosted open-source stand-ins for Slack, X, Reddit, Instagram, Spotify and
Google Drive. Its leaderboard split, \hard{} (178 tasks), asks agents to find a phone number, time or name in one app
and use it in another while an outdated or look-alike version sits nearby, or to ask the user when the request is
ambiguous. Every task is graded by code that checks the requested change and rejects any other, and every checker is
tested against executed near-miss end states. On \hard{}, the best model's 10.5\% is a mean over three tries, and no model solves more than
2.8\% of tasks on all three; the best models complete up to 89\% of the ordinary steps of multi-app requests but at most
18\% of the steps that carry a value. Fine-tuning Qwen3.5-9B on about 3.3K code-graded trajectories from our task
generator raises it from 36.3\% to 64.1\% on AndroidWorld, 8.0\% to 44.0\% on MobileWorld, 4.7\% to 22.2\% on
VenusBench-Mobile and 0\% to 7.9\% on \hard{}. Our checker tests also find that 32\% of MobileWorld GUI tasks and 14\% of audited AndroidWorld tasks accept
a wrong end state.
\end{abstract}

\IfFileExists{figures/fig_teaser.tex}{\par\vspace{2pt}\noindent\begin{minipage}{\linewidth}
\centering
\includegraphics[width=\linewidth]{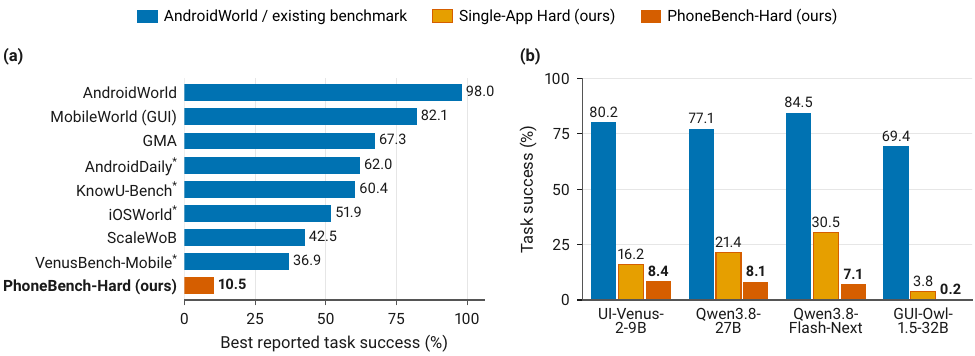}
\captionof{figure}{\textbf{\name{} is far harder than existing phone benchmarks.} (a)~Best reported task success per
benchmark (Table~\ref{tab:benchmarks}; $^{*}$graded partly by a model judge, the rest by code). \hard{}: best model,
GLM-5.3-Flash, mean over up to three tries. (b)~Our four models with AndroidWorld results: AndroidWorld (116 tasks;
developer-reported, Qwen3.8-27B measured by us over 8 runs), our Single-App Hard split (70 tasks) and \hard{} (178
tasks), both as the mean over up to three tries.}\label{fig:teaser}
\end{minipage}\par\vspace{6pt}

}{}
}]
\thispagestyle{plain}

\section{Introduction}\label{sec:intro}
Operating a phone through its screen is now a headline capability of new models, and the benchmarks used to measure
it are close to saturation. A multi-agent system reaches 98\% on AndroidWorld \citep{favreau2026multiagents,
rawles2024androidworld}, the best GUI-only entry on the MobileWorld leaderboard reaches 82\% \citep{kong2025mobileworld},
and newer suites report 37--67\%, the lowest where a model judge grades the outcome \citep{gong2026venusbench,jang2026iosworld,
sui2026androiddaily,zhu2026gma}. Read at face value, these numbers say that phone agents are nearly ready for everyday use.

They are not, for two reasons. First, \emph{benchmark tasks are mostly direct}: a typical task names the app, the item
and the change. Everyday requests are indirect. The number to text is in an email signature and was corrected later
in the thread; the meeting time in the team chat was moved; ``add the florist to my contacts'' requires finding who
the florist was. Solving such a request means finding the right value, noticing that it has been superseded, and
carrying it into another app without error. Second, \emph{the graders are untested}. Self-hosted phone benchmarks grade
with code that is assumed to be correct. When we execute deliberately wrong end states against the shipped graders,
a third of MobileWorld GUI tasks and a seventh of the AndroidWorld tasks we audited accept at least one of them
(\S\ref{sec:audit}).

\name{} is built for both. Its tasks run on the phone's own apps and on self-hosted open-source stand-ins for team
chat, a social network, a forum, photo sharing, music streaming and a cloud drive, the roles that Slack, X, Reddit,
Instagram, Spotify and Google Drive play on most phones. Figure~\ref{fig:teaser} shows the result. The four models in our panel with
AndroidWorld results score 69--85\% there but solve 4--31\% of our single-app hard tasks, and on the leaderboard
split, \hard{}, no model solves more than 10.5\% of tasks (mean over three tries). Agents use the apps well but get the facts wrong: they
complete most ordinary steps and miss the step that carries the value (\S\ref{sec:why}). Training on tasks from our
generator improves a 9B model on every benchmark we test, by 17.5 to 36.0 points on the three existing ones
(Figure~\ref{fig:training}, \S\ref{sec:training}).

\csname introteaser\endcsname
\IfFileExists{figures/fig_training.tex}{\providecommand{\colref}{gray}\providecommand{\colours}{blue}\providecommand{\Colours}{Blue}
\begin{figure}[t]
\centering
\includegraphics[width=\linewidth]{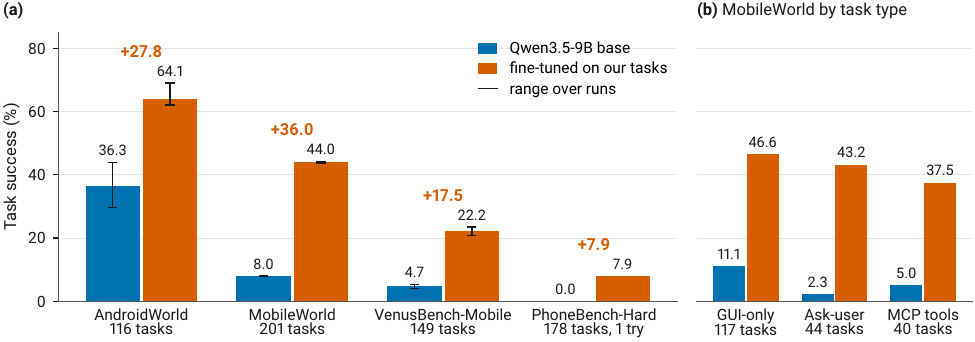}
\caption{\textbf{Fine-tuning on our generated tasks lifts a 9B model on every benchmark.} Qwen3.5-9B base (\colref)
vs.\ the same model fine-tuned on code-graded trajectories from our task generator (\colours); identical evaluation
settings, only the weights differ. (a)~Bars are the mean over runs, error bars the lowest to highest run: AndroidWorld
(116 tasks; 7 base runs, 5 fine-tuned runs), MobileWorld (201 tasks; 2 runs each), VenusBench-Mobile (149 tasks;
2 runs each) and \hard{} (178 tasks; one try each, so no error bar; 14 of 178 solved after fine-tuning). \Colours{} numbers are the absolute
gain in points. (b)~MobileWorld by task type, mean of the two runs: GUI-only (117 tasks), ask-user (44 tasks) and
MCP tools (40 tasks). With one try on \hard{}, the fine-tuned 9B model (7.9) lies between UI-Venus-2-9B (6.7) and Qwen3.8-27B (9.6),
whose one-try scores come from the leaderboard runs with different sampling.}\label{fig:training}
\end{figure}

}{}

Our contributions:
\begin{enumerate}
\item \textbf{\hard, a leaderboard of everyday multi-app requests} (178 tasks): 66 composed requests of three to five
  steps across apps, each containing a step where the right value must be found and carried, and 112 standalone
  cross-app tasks of that kind (\S\ref{sec:bench}).
\item \textbf{Graders that are tested, not assumed.} Each task's golden solution passes twice on fresh devices, all but 3 of
  the 409 executed near-miss end states counted in \S\ref{sec:audit} are rejected (those 3 only drop the last word of
  a correct message), and any
  change outside the task's allowed set fails the task.
\item \textbf{An evaluation of seven open models}, general and GUI-specialised (\S\ref{sec:results}), with a diagnosis of where they fail: agents operate the apps well but read and carry
  the wrong value (\S\ref{sec:why}).
\item \textbf{Evidence that our task generator makes useful training data.} Fine-tuning Qwen3.5-9B on about 3.3K
  successful, code-graded trajectories on tasks our generator wrote raises it from 36.3\% to
  64.1\% on AndroidWorld, from 8.0\% to 44.0\% on MobileWorld, from 4.7\% to 22.2\% on VenusBench-Mobile and from
  0\% to 7.9\% on \hard{} (\S\ref{sec:training}).
\item \textbf{An executed checker audit of AndroidWorld and MobileWorld}, task by task, with the causes of false
  accepts (\S\ref{sec:audit}).
\end{enumerate}

\section{Related work}\label{sec:related}
\paragraph{Phone benchmarks.} AndroidWorld \citep{rawles2024androidworld} introduced seeded tasks on an Android
emulator graded by code that reads device state; it is now saturated \citep{favreau2026multiagents}. MobileWorld
\citep{kong2025mobileworld} moved to self-hosted app servers with agent-user interaction and MCP tools, and GMA
\citep{zhu2026gma} adds tiers from atomic to realistic tasks on self-hosted apps. AndroidLab \citep{xu2024androidlab}
and MobileSafetyBench \citep{lee2024mobilesafetybench} also grade on emulators. A second family runs on real apps and
grades with a model judge: SPA-Bench \citep{chen2024spabench}, A3 \citep{chai2025a3}, ProBench
\citep{yang2025probench}, VenusBench-Mobile \citep{gong2026venusbench}, KnowU-Bench \citep{chen2026knowubench},
MemGUI-Bench \citep{liu2026memgui}, AndroidDaily \citep{sui2026androiddaily} and iOSWorld \citep{jang2026iosworld}.
ScaleWoB \citep{liu2026scalewob} renders synthetic apps as web pages. Table~\ref{tab:benchmarks} compares them. The
benchmarks that still leave room are mostly judged by a model; the code-graded emulator benchmarks AndroidWorld and
MobileWorld are close to saturation. \name{}
keeps code grading and self-hosting and moves the difficulty into the request itself.

\begin{table}[t]
\centering\small
\setlength{\tabcolsep}{4pt}
\caption{Interactive phone benchmarks. \emph{Best} is the best task success rate reported for the benchmark.
To our knowledge, \name{} is the only one that reports testing its checkers against executed wrong end states.}\label{tab:benchmarks}
\begin{tabular}{lrllr}
\toprule
Benchmark & Tasks & Apps & Grading & Best (\%) \\
\midrule
AndroidWorld \citep{rawles2024androidworld} & 116 & 20, emulator & code & 98.0 \\
MobileWorld \citep{kong2025mobileworld} & 201 & self-hosted servers & code & 82.1$^a$ \\
GMA \citep{zhu2026gma} & 300 & self-hosted servers & code & 67.3 \\
AndroidDaily \citep{sui2026androiddaily} & 350 & 94 real apps & model judge & 62.0 \\
KnowU-Bench \citep{chen2026knowubench} & 192 & emulator & rules + judge & 60.4 \\
iOSWorld \citep{jang2026iosworld} & 133 & 26 iOS apps & model judge & 51.9 \\
ScaleWoB \citep{liu2026scalewob} & 120 & synthetic web apps & code & 42.5 \\
VenusBench-Mobile \citep{gong2026venusbench} & 149 & 27, emulator & code + judge & 36.9 \\
\midrule
\textbf{\hard{} (ours)} & 178 & 12, self-hosted servers & tested code & 10.5 \\
\bottomrule
\end{tabular}
\\[2pt]{\footnotesize $^a$ GUI-only tasks, leaderboard entry from the model's developer; the paper's best overall result is 51.7.}
\end{table}

\paragraph{Testing the graders.} \citet{zhu2025abc} propose a checklist for outcome validity of agentic benchmarks;
\citet{dong2026misscore} audit 150 recorded failure verdicts of computer-use benchmarks and find 15\% of them wrong; SWE-ABS
\citep{yu2026sweabs} strengthens code-benchmark tests with adversarial patches and finds inflated success rates.
On desktops, OSWorld \citep{xie2024osworld} and OSWorld~2.0 \citep{yuan2026osworld2} grade with scripts. We bring
executed near-miss testing to phone benchmarks, both as an audit of existing ones and as a release gate for ours.

\paragraph{Composition, state diffs and reliability.} CompWoB \citep{furuta2023compwob} shows that web agents fail
on sequential compositions of tasks they solve alone; WorkArena++ \citep{boisvert2024workarenapp} and UI-NEXUS
\citep{guo2025uinexus} compose tasks for enterprise web apps and phones. \citet{gu2026awgen} show that RL gains on
AndroidWorld shrink from unseen instances to unseen templates and apps. AppWorld \citep{trivedi2024appworld} and
Agent-Diff \citep{pysklo2026agentdiff} grade API agents on database state diffs; we apply the same idea to the whole
phone. $\tau$-bench \citep{yao2024taubench} and \citet{gonzalez2026reliability} report repeated-run reliability.

\section{The \name{} benchmark}\label{sec:bench}
\paragraph{Environment.} Each episode runs in a Docker Android emulator connected to self-hosted servers for six
everyday roles: team chat (Mattermost, like Slack or Teams), a social network (Mastodon, like X), a forum (Lemmy, like
Reddit), photo sharing (Pixelfed, like Instagram), music streaming (Navidrome, played through the Ultrasonic client,
like Spotify) and a cloud drive (Nextcloud, like Google Drive or Dropbox). Alongside them are the phone's built-in Mail, Messages
(SMS), Calendar, Clock, Contacts, Files and Settings apps (Table~\ref{tab:apps}). Every episode starts from a fresh,
seeded device and server state with a pinned clock and logged-in accounts, so every run of a task starts from the same
content. The agent sees screenshots and acts through taps, swipes, typing, app launches and an answer action; it can
also ask the user a question, which a simulated user answers.

\begin{figure}[t]
\centering
\fbox{\begin{minipage}{0.95\linewidth}\small
\textbf{Request.} \emph{There are three things I'd like you to handle on my phone. First: Can you put the Autumn Bake Sale poster from
the Downloads folder on Mastodon under my own account? Use this as the text: ``Cakes, cookies and good causes: come
along to our autumn bake sale on the 25th!'' Second: Could you look up where my next sports massage is and email the
street address to Lucia at lucia.ferro@mailbox.example? Use the subject ``Pick-up address''. Third: Could you send
Nina what they asked for in their Mattermost message?}

\medskip
\textbf{Trap.} The third step is the value-carrying one. Nina's message reads ``What's the number of that plumber you had round?''
Two plumbers are saved in Contacts and nothing on the device shows which one came, so the agent should ask the user.

\medskip
\textbf{What the agent did.} Qwen3.8-Flash-Next posted the poster correctly, then emailed ``74 Holloway'' instead of ``74 Holloway Road'', and for
Nina picked the plumber who had once texted an unrelated holiday-closure notice and sent his saved number without
asking. The task fails on steps two and three.
\end{minipage}}
\caption{\textbf{One wrong value fails the whole request.} A three-step Composed-Hard task, verbatim, and the
episode of Qwen3.8-Flash-Next on it.}\label{fig:example}
\end{figure}
\begin{table}[t]
\centering\small
\caption{\textbf{Apps in \name.} Six roles are filled by open-source apps backed by self-hosted servers, each standing
in for a widely used commercial app; the rest are the phone's built-in apps. The cloud drive appears only in the full
benchmark; \hard{} uses the other twelve apps.}\label{tab:apps}
\begin{tabular}{lll}
\toprule
Everyday role & App in \name{} & Commercial counterpart \\
\midrule
Team chat & Mattermost & Slack, Microsoft Teams \\
Social network & Mastodon & X (Twitter) \\
Forum & Lemmy (Jerboa client) & Reddit \\
Photo sharing & Pixelfed (PixelDroid client) & Instagram \\
Music streaming & Navidrome (Ultrasonic client) & Spotify \\
Cloud drive & Nextcloud & Google Drive, Dropbox \\
\midrule
Mail & built-in Mail app & Gmail \\
Messages, Calendar, Clock, Contacts, & built-in Android apps & the same apps on any phone \\
\quad Files, Settings & & \\
\bottomrule
\end{tabular}
\end{table}

\paragraph{Tasks.} Every task is an everyday request in the user's own words. The hard tasks share one idea: the phone
holds the right value and a believable outdated or look-alike one. An exact value (a phone number, a time, an amount, a
name) must be found in one app and used in another, and the environment also contains a plausible wrong version of it,
such as an older number in a mail thread, a meeting time that was moved later in the chat, or two contacts who both
match the request. Some tasks can only be solved by asking the user. Composed tasks join three to five steps on
different apps into one request; exactly one step is of the hard kind and the others are ordinary chores.

\paragraph{Coverage.} Table~\ref{tab:categories} lists what \hard{} asks for. Each of its seven categories is an
errand people run on their phones daily: passing on a number someone asked for, saving a person found on social media
as a contact, telling a friend when an event starts, setting an alarm for a pickup, collecting money after a trip,
making a playlist a friend asked for, and asking the user when the phone cannot settle the request (39 tasks). Asking
is not free elsewhere: when the answer is on the phone, a question to the user fails the task. Each category touches
two to four apps. Around the hard step, the 66 composed requests chain 200 ordinary steps across the same apps, from
posting with an image on the social network and pinning a team-chat message to starring songs, zipping files, adding
calendar events and changing settings. \hard{} uses twelve apps; the full benchmark adds the cloud drive.

\begin{table}[!t]
\centering\small
\setlength{\tabcolsep}{3.5pt}
\renewcommand{\arraystretch}{1.15}
\caption{\textbf{What \hard{} asks for.} Top: the seven hard categories, a shortened example request for each (people,
events and values change from task to task), the apps involved (count in brackets) and the trap. In every hard task
the phone holds the right value and a believable outdated or look-alike one. In \emph{Ask the user} the phone cannot
settle the request and the agent must ask before acting; in the other categories asking is itself a failure.
Bottom: the ordinary chores that composed requests chain around the hard step. C-H: Composed-Hard requests (by hard
step) or ordinary steps; X: standalone Cross-App tasks.}\label{tab:categories}
\begin{tabular}{>{\raggedright\arraybackslash}p{2.5cm}>{\raggedright\arraybackslash}p{4.2cm}>{\raggedright\arraybackslash}p{3.3cm}>{\raggedright\arraybackslash}p{3.9cm}rr}
\toprule
Category & Example request & Apps & Trap & C-H & X \\
\midrule
Share a contact & ``Gareth emailed asking for a number. Reply to him.''
 & Mail or team chat, Contacts, Messages (3) & New number texted later; look-alike of another trade & 20 & 20 \\
Ask the user & ``Priya emailed wanting a contact number. Sort out the reply.''
 & Mail or team chat, Contacts, Messages (3) & Two people fit; ask, then send the new number & 19 & 20 \\
Save a social contact & ``Add the florist Milo tagged on Pixelfed to my contacts.''
 & Photo sharing, Contacts (2) & Later comment swaps the credit; old and new number on profile & 15 & 20 \\
Send an event time & ``Let Marcus know when Saturday's concert starts.''
 & Calendar, Mail, Messages (3) & Start moved after the calendar entry & -- & 20 \\
Set a pickup alarm & ``Alarm an hour before my taxi tomorrow.''
 & Mail, Messages, Clock (3) & Pickup moved by text; flight time nearby & -- & 20 \\
Settle trip money & ``Chase up the Lisbon money; tell me the total owed.''
 & Team chat, Mail, Messages, Contacts (4) & ``Paid!'' vs.\ bank emails: bounced, partial, unannounced & 5 & 5 \\
Build a playlist & ``Make the playlist Ana texted about; tell her when done.''
 & Messages, music streaming (2) & Rules change in later texts; live and acoustic look-alikes & 7 & 7 \\
\midrule
Total hard & & & & 66 & 112 \\
\midrule
\multicolumn{2}{l}{\textit{Ordinary chores in composed requests}} & Apps & Example chores & C-H & \\
\midrule
Social network & & Mastodon, Files & Post with an image; unfollow by bio; edit profile & 46 & \\
Music & & Navidrome & Star songs by a rule & 38 & \\
Settings & & Settings & Brightness, Do Not Disturb, Bluetooth, torch & 37 & \\
Team chat & & Mattermost & Pin a decision; edit a header; message volunteers & 35 & \\
Files & & Files & Zip and clean up files & 20 & \\
Calendar and mail & & Calendar, Mail & Add an event; email an address & 19 & \\
Forum & & Lemmy & Look up a post & 5 & \\
\midrule
Total ordinary & & & & 200 & \\
\bottomrule
\end{tabular}
\end{table}

\paragraph{Example.} Figure~\ref{fig:example} shows a three-step Composed-Hard request, verbatim, and how an agent
fails it.

\paragraph{Grading.} A task passes only if (i) every requested change is present, (ii) a diff of the whole device and
server state shows no change outside the task's allowed set, (iii) any requested answer matches after normalising
numbers, times, dates and units, and (iv) the agent stays within the step budget (50 steps per composed step, 80 for
standalone cross-app tasks). Only the answer action counts as the agent's answer.
The headline metric is this strict success rate. As a secondary metric we report partial credit: a composed request
scores the fraction of its steps that pass their checks, a request whose steps all pass but that fails a request-wide
check (collateral change, step budget, call format) scores as if one step had failed, and single-step tasks score as
strict.

\paragraph{Release gates.} A task is released only if its golden solution passes twice on fresh devices with no change
outside the allowed set, and its checker is run against executed near-miss end states that a user would call wrong; of
the 409 counted in \S\ref{sec:audit}, all but 3 are rejected on the final checkers, and those 3 only drop the last
word of a correct message. Task difficulty was set on separate development instances; the benchmark instances are fresh.
Every category of \hard{} is solved end to end by at least one model in our runs, asking the user in three episodes,
all on one request.

\paragraph{Splits.} \hard{} (178 tasks) is the leaderboard. The full benchmark (338 tasks) adds two easier splits
(Table~\ref{tab:splits}): 90 composed requests whose hard step stays inside one or two apps, and 70 single-app hard
tasks. Results on all splits are in Appendix~\ref{app:full}.

\begin{table}[t]
\centering\small
\caption{\textbf{\name{} splits.} The two \hard{} splits share the hard categories of Table~\ref{tab:categories}. The
Composed split's hard steps are simpler: answering with a code found in mail, emailing a file from Downloads, replying
to an order after asking the user, or voting in a social-network poll as agreed in team chat. Single-App Hard mostly
stays within one app, such as replying to an order in team chat or answering a text with a detail, sometimes after
asking the user, plus cloud-drive tasks.}\label{tab:splits}
\begin{tabular}{llrr}
\toprule
Split & Content & Tasks & Step budget \\
\midrule
\multicolumn{4}{l}{\emph{\hard{} (leaderboard)}} \\
Composed-Hard & 3--5 step requests with one cross-app hard step & 66 & 150--250 \\
Cross-App & standalone cross-app hard tasks & 112 & 80 \\
\midrule
\multicolumn{4}{l}{\emph{Full benchmark only}} \\
Composed & 3--5 step requests with a simpler hard step & 90 & 150--250 \\
Single-App Hard & hard tasks within one app, incl.\ cloud drive & 70 & 50--80 \\
\midrule
Total & & 338 & \\
\bottomrule
\end{tabular}
\end{table}

\section{Results}\label{sec:results}
\paragraph{Models.} We evaluate seven open models, all served on our own cluster with vLLM: four general multimodal models
(GLM-5.3-Flash \citep{card_glm53flash}, Inkling-2-Small \citep{card_inkling2}, Qwen3.8-Flash-Next \citep{card_flashnext}
and DeepSeek-V4.1-Flash \citep{card_deepseekv41}), the dense Qwen3.8-27B, and two models trained specifically for GUI
agents, UI-Venus-2-9B \citep{card_uivenus2} and GUI-Owl-1.5-32B \citep{card_guiowl15}.

\paragraph{Agent.} All models get the same system prompt, tool schema and history format. At each step the model
receives the request, the current screenshot, up to 19 earlier screenshots (beyond that the oldest ten are replaced
by a short text note) and its earlier actions, reasons with thinking enabled, and emits one action: tap, long press, swipe, type, press a key or system
button, open an app, wait, ask the user, answer, or end the episode. Per-model differences are limited to a thin
adapter for the model's coordinate convention and tool-call syntax (for two models, one added sentence in the system
prompt states the coordinate convention), the thinking budget (4,096 tokens for Qwen3.8-27B, 16,384 for the others)
and maximum reasoning effort for GLM-5.3-Flash. Open models are sampled at temperature 1.0 and top-$p$ 0.95 with a
seed derived from the task, try and step, so runs are reproducible. Every model gets three tries per task on \hard{}; we report pass@1, the success rate averaged
over the three tries \citep{chen2021codex}, as the headline, with the share of tasks solved in at least one try (pass@3) and in
all three (pass$^3$~\citep{yao2024taubench}). Appendix~\ref{app:prompt} describes the harness.

\begin{table}[t]
\centering\footnotesize
\caption{\textbf{\hard{} leaderboard}: task success rate (\%) as pass@1, the mean over three tries per task (successes/graded tries
in brackets). A task counts only if every requested change is present, nothing else on the device changed, and the
agent stayed within the step budget. \emph{Total} is the headline. pass@3: share of tasks solved in at least one of
the three tries; pass$^3$: share solved in all three (over the 176--178 tasks with three graded tries). Partial
credit: a composed request scores the fraction of its steps completed (a request whose steps are all done but that
fails a request-wide check scores as if one step failed); single-step Cross-App tasks score as strict.
$^g$\,Trained for GUI agents.}\label{tab:leaderboard}
\setlength{\tabcolsep}{2.6pt}
\begin{tabular}{lccccccc}
\toprule
 & \multicolumn{7}{c}{\hard{} (178 tasks)} \\
\cmidrule(lr){2-8}
 & \multicolumn{3}{c}{pass@1 (mean of 3 tries)} & \multicolumn{2}{c}{Over 3 tries} & \multicolumn{2}{c}{Partial credit} \\
\cmidrule(lr){2-4}\cmidrule(lr){5-6}\cmidrule(lr){7-8}
Model & Composed-Hard (66) & Cross-App (112) & \textbf{Total (178)} & pass@3 & pass$^3$ & Composed & Total \\
\midrule
GLM-5.3-Flash \citep{card_glm53flash} & 10.1 \tiny{(20/198)} & 10.7 \tiny{(36/335)} & \textbf{10.5} \tiny{(56/533)} & 22.6 & 1.1 & 67.7 & 31.9 \\
Inkling-2-Small \citep{card_inkling2} & \phantom{0}6.1 \tiny{(12/198)} & 12.8 \tiny{(43/336)} & \textbf{10.3} \tiny{(55/534)} & 20.2 & 1.7 & 61.5 & 30.8 \\
UI-Venus-2-9B$^g$ \citep{card_uivenus2} & \phantom{0}4.0 \tiny{(8/198)} & 11.0 \tiny{(37/336)} & \phantom{0}\textbf{8.4} \tiny{(45/534)} & 19.1 & 0.6 & 55.5 & 27.5 \\
Qwen3.8-27B & \phantom{0}2.5 \tiny{(5/198)} & 11.3 \tiny{(38/335)} & \phantom{0}\textbf{8.1} \tiny{(43/533)} & 16.4 & 1.1 & 61.5 & 30.0 \\
Qwen3.8-Flash-Next \citep{card_flashnext} & \phantom{0}4.0 \tiny{(8/198)} & \phantom{0}8.9 \tiny{(30/336)} & \phantom{0}\textbf{7.1} \tiny{(38/534)} & 14.6 & 1.1 & 68.1 & 30.9 \\
DeepSeek-V4.1-Flash \citep{card_deepseekv41} & \phantom{0}0.5 \tiny{(1/198)} & \phantom{0}9.8 \tiny{(33/336)} & \phantom{0}\textbf{6.4} \tiny{(34/534)} & 10.7 & 2.8 & 44.7 & 22.8 \\
GUI-Owl-1.5-32B$^g$ \citep{card_guiowl15} & \phantom{0}0.0 \tiny{(0/198)} & \phantom{0}0.3 \tiny{(1/334)} & \phantom{0}\textbf{0.2} \tiny{(1/532)} & \phantom{0}0.6 & 0.0 & 17.8 & \phantom{0}6.8 \\
\bottomrule
\end{tabular}
\end{table}

\paragraph{No model solves \name.} Table~\ref{tab:leaderboard} reports \hard. The best model, GLM-5.3-Flash, solves 10.5\%
of tasks over three tries: 10.1\% of the composed requests and 10.7\% of the standalone cross-app tasks. Inkling-2-Small
solves 10.3\%, UI-Venus-2-9B, trained for GUI agents, 8.4\%, Qwen3.8-27B 8.1\%, Qwen3.8-Flash-Next 7.1\% and
DeepSeek-V4.1-Flash 6.4\%; GUI-Owl-1.5-32B, also trained for GUI agents, succeeds once in 532 tries. A single
model's success rate moves by up to three points from one try to the next, so gaps of that size between models are not
meaningful. Every model is graded on all 178 tasks.
Success is rarely repeatable: the best models solve about a fifth of the tasks at least once in three tries
(pass@3 up to 22.6\%), but no model solves more than 2.8\% of the tasks in all three tries (pass$^3$; Figure~\ref{fig:tries}).
Partial credit, which counts the completed steps of a composed request,
puts the models other than GUI-Owl-1.5-32B at 45--68\% on Composed-Hard, against 0.5--10.1\% strict: they
complete most of each request and miss the step that carries the value (Section~\ref{sec:why}).

\IfFileExists{figures/fig_tries.tex}{\begin{figure}[t]
\centering
\includegraphics[width=\linewidth]{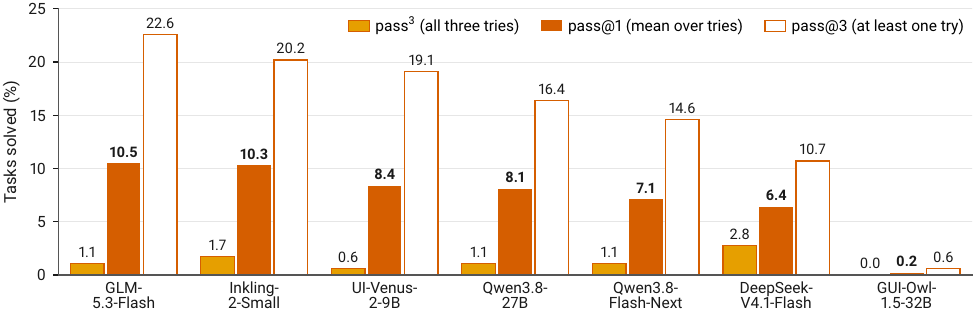}
\caption{\textbf{About a fifth of \hard{} tasks are solved at least once, but almost none every time.} Per model,
up to three tries on each of the 178 \hard{} tasks: pass$^3$, the share of tasks solved in all three tries (light),
pass@1, the mean success over tries (solid; the headline), and pass@3, the share solved in at least one try (outline). Models are sorted by pass@1.
No model solves more than 2.8\% of the tasks in all three tries.}\label{fig:tries}
\end{figure}

}{}

\paragraph{Asking is not enough.} Of the 817 episodes on the 39 tasks that require asking the user (three tries each),
three succeed, all on the same composed request: GLM-5.3-Flash once and Inkling-2-Small twice found the person's new
number, asked which of two matching people was meant, and sent it. Qwen3.8-27B and Qwen3.8-Flash-Next ask which person
was meant in most of the standalone ask episodes (54 of 59 and 40 of 60), yet 49 of the 54 and 39 of the 40 give an
outdated or wrong number: the question settles who was meant, not which of that person's numbers is current.

\paragraph{The same models do well elsewhere.} Table~\ref{tab:samemodel} and Figure~\ref{fig:teaser} compare the
same models across benchmarks. Models that score 69--85\% on AndroidWorld solve 4--31\% of \name's single-app hard
tasks: every model drops, so the drop is a property of the tasks, not of one model. On these single-app tasks, GUI
specialisation did not help.
UI-Venus-2-9B and GUI-Owl-1.5-32B, both trained for phone agents, solve 16\% and 4\% of single-app hard tasks, below
both general Qwen3.8 models (30.5\% and 21.4\%). On \hard{}, UI-Venus-2-9B ranks third (8.4\%), while GUI-Owl-1.5-32B succeeds once in 532 tries: its taps land where intended and
it completes about a quarter of the ordinary steps of composed requests (141 of 600), but when the value must be found on the
device it most often types one it never saw (6 of 15 failed episodes we read), or carries a wrong one (3 of 15).

\begin{table}[t]
\centering\small
\setlength{\tabcolsep}{4.5pt}
\caption{\textbf{Same models, different benchmarks} (task success rate, \%). AndroidWorld and MobileWorld (GUI-only)
as reported on each model card, except Qwen3.8-27B ($^m$: measured by us, mean of 8 runs). \name{}: mean over up to 3 tries per task; rows in \hard{} order.}\label{tab:samemodel}
\begin{tabular}{lccccc}
\toprule
 & \multicolumn{2}{c}{Existing benchmarks} & \multicolumn{3}{c}{\name{} (ours)} \\
\cmidrule(lr){2-3}\cmidrule(lr){4-6}
Model & AndroidWorld & MobileWorld & Composed & Single-App Hard & \hard \\
\midrule
UI-Venus-2-9B & 80.2 & 65.8 & 11.9 & 16.2 & \phantom{0}8.4 \\
Qwen3.8-27B & 77.1$^m$ & 82.0$^m$ & 17.8 & 21.4 & \phantom{0}8.1 \\
Qwen3.8-Flash-Next & 84.5 & -- & 30.0 & 30.5 & \phantom{0}7.1 \\
GUI-Owl-1.5-32B & 69.4 & 46.8 & \phantom{0}0.0 & \phantom{0}3.8 & \phantom{0}0.2 \\
\bottomrule
\end{tabular}
\end{table}

\paragraph{Harder than other phone benchmarks.} The best reported results on existing interactive phone benchmarks
range from 37\% to 98\% (Table~\ref{tab:benchmarks}); the lowest of them, VenusBench-Mobile, is graded partly by a
model judge. On \hard, graded only by code, the best result is 10.5\%.

\section{Training on generated tasks}\label{sec:training}
The tasks that make \name{} hard come from a task generator that writes everyday requests together with code
checkers. The same generator can produce training data. We test this with a small open model.

\paragraph{Setup.} We fine-tune Qwen3.5-9B (full fine-tune, one epoch) on 3,313 successful, code-graded trajectories
on tasks from our generator. The tasks are our own, not benchmark tasks, and run
on the open-source apps that the MobileWorld and AndroidWorld environments use.

\paragraph{Evaluation.} Base and trained models use the same agent format, step budgets, sampling and serving; only
the weights differ. Both run at temperature 0 with thinking on. We evaluate on AndroidWorld (116 tasks), MobileWorld
(201 tasks), the 149 primary tasks of VenusBench-Mobile and \hard{} (178 tasks, one try). Unlike the leaderboard in
Section~\ref{sec:results}, which samples at temperature 1.0 over three tries, \hard{} gets one try per task here. Runs in which an emulator failed permanently mid-run were discarded (three on AndroidWorld).

\paragraph{Results.} Training on generated tasks improves every benchmark (Figure~\ref{fig:training}). On
AndroidWorld the trained model scores 64.1\% (mean of 5 runs, s.d.\ 2.8) against 36.3\% for the base model (mean of 7
runs, s.d.\ 4.9). On MobileWorld it rises from 8.0\% to 44.0\% (2 runs each): from 11.1\% to 46.6\% on GUI-only tasks,
from 2.3\% to 43.2\% on tasks that require asking the user and from 5.0\% to 37.5\% on tasks with MCP tools. Two
Mattermost tasks never initialise in our environment for either model; the scores are over all 201 tasks. On
VenusBench-Mobile it rises from 4.7\% to 22.2\% (2 runs each).

On \hard{} the base model solves no task (0 of 178) and the trained model solves 14 (7.9\%), all of them standalone
Cross-App tasks (14 of 112; 0 of 66 Composed-Hard); partial credit rises from 1.8\% to 28.1\%. The trained 9B model's
one-try score is close to that of UI-Venus-2-9B (6.7\%) and below that of Qwen3.8-27B (9.6\%) on their first try
(Appendix~\ref{app:full}), although those models were sampled at temperature 1.0.

\paragraph{Decontamination and filtering.} The pool holds successful, code-graded tries on 5,245 generated tasks;
tasks that are themselves in \name{} were never in it. Decontamination ran first. Every task that shares its task type
and random seed with a \name{} task, or with a step of a composed \name{} request, was removed (22 tasks), so no
\hard{} task is in training and no training task is an instance of a Composed-Hard request type. A near-copy check
then compared every training request with 868 task goals and templates from AndroidWorld, MobileWorld,
VenusBench-Mobile and GMA: after names, numbers, dates, quoted values and template slots are normalised to one
placeholder, a training task is dropped if it shares any 13-word sequence with a benchmark goal (benchmark goals of
6--12 words must appear whole). It found 0 near-copies; on 40 planted, altered copies per benchmark it recovers 36--39,
missing only short templates. The check covers plain-text goals of at least six words; 54 goals in other scripts (53 of them
Chinese MobileWorld goals) and 21 shorter goals are outside it. Our training requests are in English. Tasks that are similar to benchmark tasks but not copies are kept on purpose. Episode filtering then reduced 12,688
successful episodes to 3,415: automatic checks (well-formed actions, an explicit final action, no loops or overlong
reasoning, no test-aware wording or environment addresses) removed 3,489; rules from hand audits (outdated task
versions, broken seeds, real brands or public figures, needless questions to the user, non-Latin output) removed
1,609; we kept one episode per task (3,422); and a check that every record matches exactly what the model saw dropped
7. The 3,415 episodes are split into 3,313 for training and 102 for validation.

\paragraph{The gain is not memorisation.} Thirteen of the trained model's 14 \hard{} passes come from chores it
never saw in training. Twelve are on setting an alarm before a pickup whose time the taxi firm later moved by text (12 of 20
tasks), and one is on saving a supplier credited in a photo-sharing post as a contact (1 of 20); training contains no
episodes of either chore. The three cross-app chores that do have training episodes, on other task instances, score 1
of 20, 0 of 20 and 0 of 7, and the remaining two score none.

\paragraph{Caveat.} The base model is evaluated in our agent format, which it was not trained on, so part of the gain
on every benchmark may be adaptation to the format, including the ask action on MobileWorld's ask-user tasks. The
\hard{} passes on chores absent from training show that the gain is not memorisation of task types, but they do not
separate new skill from format adaptation.

\section{Why agents fail}\label{sec:why}
\paragraph{Agents can operate the apps.} Inside a Composed-Hard request, the ordinary steps (editing a channel header,
starring songs, zipping files, replying on the social network with a time from the calendar) are completed 79--89\%
of the time by Qwen3.8-27B, Qwen3.8-Flash-Next, GLM-5.3-Flash and Inkling-2-Small, but the value-carrying step of
the same request only 6--18\% of the time (Figure~\ref{fig:steps}, three tries per request; per-chore counts in
Appendix~\ref{app:full}). Pooled over the seven
models, an agent emails back a tradesperson's current number in 5\% of attempts, answers a colleague's request for a
number in 8\%, saves the supplier credited in a photo in 5\% and reminds the right friends of the right amounts after
a trip in 8\%. The easiest value-carrying step, a playlist built from rules spread across several texts, succeeds in 24\%.

\IfFileExists{figures/fig_steps.tex}{\providecommand{\colref}{gray}\providecommand{\colours}{blue}\providecommand{\Colours}{Blue}
\begin{figure}[t]
\centering
\includegraphics[width=\linewidth]{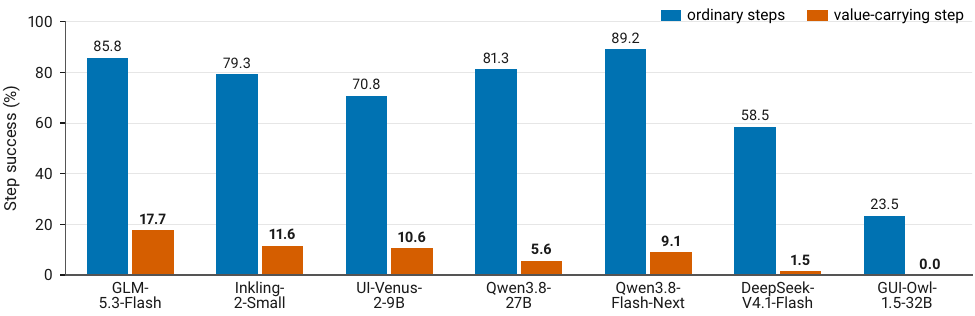}
\caption{\textbf{Agents do the chores and miss the value.} Inside the 66 Composed-Hard requests, three tries each:
success rate on the ordinary steps (\colref; 200 steps, 600 step attempts per model) vs.\ on the one value-carrying
step of each request (\colours; 198 step attempts per model). Models in the same order as
Figure~\ref{fig:tries}.}\label{fig:steps}
\end{figure}

}{}

\paragraph{They read the wrong value.} To separate reading from acting, we built two-app chains where a value read in
one app must be used in another and a plausible wrong value sits at the source, and ran each half alone as well. On
50 chains, Qwen3.8-27B passes 4\%, Qwen3.8-Flash-Next 6\% and GLM-5.3-Flash 14\%. The same models complete the second
step alone 84\%, 100\% and 40\% of the time, but read the source correctly alone only 14\%, 32\% and 20\% of the time,
and in the chains, pooled over the three models, 54\% of the wrong source reads (69 of 128) carry the planted wrong value forward.
The failure is in finding and checking the value, not in operating the
second app.

\paragraph{They ask too rarely, and asking is not enough.} When a request names a person who matches two contacts,
on the standalone tasks Inkling-2-Small and UI-Venus-2-9B almost never ask (2 of 60 episodes each) and GLM-5.3-Flash and
DeepSeek-V4.1-Flash ask in under half of the cases (18 of 59 and 22 of 60); otherwise they pick one and act. In the example
of Section~\ref{sec:bench}, Qwen3.8-Flash-Next sent the number of the plumber who had once texted the user. Models that do
ask mostly still fail: the question settles who was meant, but they then send a superseded or wrong number for that
person; three of the 817 must-ask episodes succeed (Section~\ref{sec:results}).

\paragraph{Protocol.} Some models compute the right answer and never submit it: in some episodes Qwen3.8-Flash-Next states the
correct total in its reasoning without calling the answer action. We count this as a failure, as a user would.

\paragraph{Collateral changes are rare.} Of the Composed-split episodes that completed every requested step, 2 of 281
(0.7\%) changed something else on the device: one starred a contact and one marked a chat channel as a favourite. Agents fail by not doing the right thing, not by
doing damage.

\section{How reliable are phone-benchmark graders?}\label{sec:audit}
For all 117 MobileWorld GUI tasks and the 91 AndroidWorld tasks in our audit we generated candidate wrong and right end states from the task and
its grader and ran each against the benchmark's own grader: live for MobileWorld, against a mocked device layer or
by direct calls for AndroidWorld, and by reading the grader where no harness could run it (Table~\ref{tab:audit}). 32\% of MobileWorld GUI tasks and
14\% of these AndroidWorld tasks accept at least one wrong end state; these are lower bounds, since tasks
we could not test count as clean. The causes are mundane: changes outside the asked-for item are ignored, a requested field is never compared, a substring match accepts a wrong value (for example an SMS reply passes
if it merely contains ``ok'', so ``booking'' passes), and presence is checked instead of content.

The same test is a release gate for \name. On the final checkers, none of 227 executed near-misses on composed tasks is
accepted, 2 of 125 on single-app hard tasks and 1 of 57 on the cross-app tasks; all three accepted cases only drop
the last word of an otherwise correct message.

\begin{table}[t]
\centering\small
\caption{\textbf{Existing graders accept wrong end states; ours almost never do.} Executed checker audit, rate in \% (count in brackets). Existing benchmarks: a task counts if any tested wrong end state is accepted, or any
tested right end state rejected. \name: executed near-miss and alternative-correct end states.}\label{tab:audit}
\begin{tabular}{lcc}
\toprule
Checkers & Accept a wrong state (\%) & Reject a right state (\%) \\
\midrule
MobileWorld, GUI tasks & 32.5 \tiny{(38/117)} & 6.0 \tiny{(7/117)} \\
MobileWorld, non-GUI tasks & 20.2 \tiny{(17/84)} & 6.0 \tiny{(5/84)} \\
AndroidWorld & 14.3 \tiny{(13/91)} & 0.0 \tiny{(0/91)} \\
\midrule
\name, composed & 0.0 \tiny{(0/227)} & 0.0 \tiny{(0/7)} \\
\name, single-app hard & 1.6 \tiny{(2/125)} & -- \\
\name, cross-app tasks & 1.8 \tiny{(1/57)} & 0.0 \tiny{(0/4)} \\
\bottomrule
\end{tabular}
\end{table}

\section{Limitations}\label{sec:limits}
\paragraph{Stand-in apps.} Tasks run on self-hosted open-source stand-ins for widely used apps (team chat, social
network, forum, photo sharing, music, cloud drive) and on the stock Android apps. This makes every episode
reproducible and every outcome checkable from server and device state, but it cannot capture the look and churn of
the commercial apps people use.

\paragraph{Coverage of models and tries.} The leaderboard covers open models only. Every model gets three tries per
task on \hard{}; on the appendix splits, DeepSeek-V4.1-Flash and 20 of the 70 Single-App Hard tasks got one try.

\paragraph{Training study.} The training result is for one model size, one data set and one epoch. Base and trained
models were evaluated in our agent format, which the base model was not trained on, so part of the gain on every
benchmark may be format adaptation (\S\ref{sec:training}). The trained model was run once per task on
\hard{}, at temperature 0.

\paragraph{Grader tests are lower bounds for other benchmarks.} Our audit of existing graders counts a task as
affected only when an executed wrong end state is accepted; graders we could not execute count as clean.

\paragraph{Human baseline.} A human baseline on a sample of \hard{} is left to future work.

\section{Conclusion}\label{sec:conclusion}
The best model we test solves 10.5\% of \hard{} over three tries, and the four models with AndroidWorld results,
69--85\% there, solve at most 8.4\%; no model solves more than 2.8\% of the tasks on all three tries. They complete most ordinary steps of a request and miss the step where the
right value must be found among a believable wrong one, and asking rarely rescues them: 3 of 817 episodes that
require a question to the user succeed. Every task is graded by
tested code, and the same tests show that a third of MobileWorld GUI tasks and a seventh of the audited AndroidWorld
tasks accept a wrong end state. Tasks from our generator also make effective training data: one epoch on about 3.3K
trajectories lifts Qwen3.5-9B from 36.3\% to 64.1\% on AndroidWorld, from 8.0\% to 44.0\% on MobileWorld, from 4.7\% to
22.2\% on VenusBench-Mobile and from 0\% to 7.9\% on \hard{}, where 13 of its 14 passes come from chores absent from training.

\bibliographystyle{plainnat}
\bibliography{references}

\clearpage
\appendix
\section{Additional results}\label{app:full}
\paragraph{Step-level results.} Table~\ref{tab:parts} breaks down the Composed-Hard results of
Section~\ref{sec:why} by step: the value-carrying step of each request, by hard category, and all ordinary steps.

\begin{table}[!ht]
\centering\small
\setlength{\tabcolsep}{3.5pt}
\caption{\textbf{Step-level results inside Composed-Hard}, three tries per request. Each request has one value-carrying step, which contains a plausible wrong value, and two to four ordinary
steps. Value-carrying steps by hard category (successes/attempts; the first two columns pool Share a contact and Ask the
user by the channel of the request), then the success rate (\%) over all value-carrying and
all ordinary steps.}\label{tab:parts}
\begin{tabular}{lccccccc}
\toprule
 & \multicolumn{5}{c}{Value-carrying step, by category} & \multicolumn{2}{c}{All steps of the kind} \\
\cmidrule(lr){2-6}\cmidrule(lr){7-8}
Model & \shortstack{Share/ask\\(mail)} & \shortstack{Share/ask\\(chat)} & \shortstack{Save social\\contact} & \shortstack{Settle trip\\money} & \shortstack{Build a\\playlist} & Value-carrying & Ordinary \\
\midrule
GLM-5.3-Flash & 9/57 & 13/60 & 1/45 & 1/15 & 11/21 & 17.7 \tiny{(35/198)} & 85.8 \tiny{(515/600)} \\
Inkling-2-Small & 4/57 & 1/60 & 12/45 & 1/15 & 5/21 & 11.6 \tiny{(23/198)} & 79.3 \tiny{(476/600)} \\
UI-Venus-2-9B & 5/57 & 12/60 & 2/45 & 1/15 & 1/21 & 10.6 \tiny{(21/198)} & 70.8 \tiny{(425/600)} \\
Qwen3.8-27B & 0/57 & 1/60 & 0/45 & 3/15 & 7/21 & \phantom{0}5.6 \tiny{(11/198)} & 81.3 \tiny{(488/600)} \\
Qwen3.8-Flash-Next & 0/57 & 6/60 & 0/45 & 1/15 & 11/21 & \phantom{0}9.1 \tiny{(18/198)} & 89.2 \tiny{(535/600)} \\
DeepSeek-V4.1-Flash & 1/57 & 0/60 & 1/45 & 1/15 & 0/21 & \phantom{0}1.5 \tiny{(3/198)} & 58.5 \tiny{(351/600)} \\
GUI-Owl-1.5-32B & 0/57 & 0/60 & 0/45 & 0/15 & 0/21 & \phantom{0}0.0 \tiny{(0/198)} & 23.5 \tiny{(141/600)} \\
\bottomrule
\end{tabular}
\end{table}

\paragraph{All splits.} Table~\ref{tab:full} reports every model on every split. The two appendix splits, which
are easier than the leaderboard, were run with up to three tries per task; we report the mean per-task success rate. Their tasks are
fresh instances, separate from those used during development.

\begin{table}[!ht]
\centering\small
\caption{\textbf{All splits}, task success rate (\%), mean over up to three tries per task (one try for
DeepSeek-V4.1-Flash on Composed and Single-App Hard, and for 20 of the 70 Single-App Hard tasks). \emph{Total, try 1}:
\hard{} on the first try alone. \emph{Full}: all 338 tasks.}\label{tab:full}
\setlength{\tabcolsep}{4pt}
\begin{tabular}{lccccccc}
\toprule
 & \multicolumn{2}{c}{Appendix splits} & \multicolumn{4}{c}{\hard} & \\
\cmidrule(lr){2-3}\cmidrule(lr){4-7}
Model & Composed & Single-App Hard & Composed-Hard & Cross-App & \textbf{Total} & Total, try 1 & \textbf{Full} \\
 & (90) & (70) & (66) & (112) & (178) & (178) & (338) \\
\midrule
GLM-5.3-Flash & 46.7 & 42.4 & 10.1 & 10.7 & 10.5 & 10.7 & 26.7 \\
Inkling-2-Small & 28.5 & 51.9 & \phantom{0}6.1 & 12.8 & 10.3 & \phantom{0}8.4 & 23.8 \\
UI-Venus-2-9B & 11.9 & 16.2 & \phantom{0}4.0 & 11.0 & \phantom{0}8.4 & \phantom{0}6.7 & 10.9 \\
Qwen3.8-27B & 17.8 & 21.4 & \phantom{0}2.5 & 11.3 & \phantom{0}8.1 & \phantom{0}9.6 & 13.4 \\
Qwen3.8-Flash-Next & 30.0 & 30.5 & \phantom{0}4.0 & \phantom{0}8.9 & \phantom{0}7.1 & \phantom{0}9.0 & 18.0 \\
DeepSeek-V4.1-Flash & 23.3 & 21.4 & \phantom{0}0.5 & \phantom{0}9.8 & \phantom{0}6.4 & \phantom{0}6.7 & 14.0 \\
GUI-Owl-1.5-32B & \phantom{0}0.0 & \phantom{0}3.8 & \phantom{0}0.0 & \phantom{0}0.3 & \phantom{0}0.2 & \phantom{0}0.0 & \phantom{0}0.9 \\
\bottomrule
\end{tabular}
\end{table}

\section{Task categories}\label{app:families}
\hard{} contains 178 tasks: 66 composed requests and 112 standalone Cross-App tasks, each in one of the seven
categories of Table~\ref{tab:categories}. In every category the phone holds the right value and a believable outdated
or look-alike one, and the checker reads the end state of the device and servers. Every task also fails on any change
outside the set of changes it allows (an extra contact, message or alarm, an edited event) and on exceeding its step
budget.

\paragraph{Composed requests (66).} One message asks for three, four or five things (21, 22 and 23 requests), with a
budget of 50 actions per requested step. Exactly one step belongs to a hard category (Share a contact or Ask the user
39, Save a social contact 15, Build a playlist 7, Settle trip money 5); the others are ordinary chores on the same
phone, from posting on the social network to changing a quick setting (Table~\ref{tab:categories}, bottom). The
request passes only if every step passes.

\paragraph{Share a contact and Ask the user (40 standalone, 39 composed).} Someone asks, by e-mail or in team chat,
for the number of a tradesperson the user has used, and the agent must reply where they asked. The number on the phone
has since changed. In Ask the user (20 standalone, 19 composed) two people of the same trade fit and nothing on the
phone says which one was meant, so the agent must ask the user first; in the other tasks a question to the user fails.
The reply must contain the current number and must not give an outdated or wrong one.

\paragraph{Save a social contact (20 standalone, 15 composed).} A friend's photo-sharing post credits the people
behind an event; the agent must add one of them to Contacts with the number from their profile. The credit or the
number may have been corrected, and look-alike accounts exist. Exactly one new contact with the right person and
number must be created.

\paragraph{Send an event time (20 standalone).} A friend asks by text when an event starts; the start time has
changed since it was put in the calendar. The reply must give the current start time and no superseded time.

\paragraph{Set a pickup alarm (20 standalone).} The agent must set an alarm a given time before a booked pickup. The
pickup time has changed since the booking, and other times are nearby. Exactly one new alarm at the right time must
be created.

\paragraph{Settle trip money (5 standalone, 5 composed).} The agent must remind every friend who still owes money for
a group trip of the amount they owe and report the total outstanding. Who has paid must be worked out from the group
chat and the bank's e-mails, which do not always agree. Everyone who owes, and nobody else, must be messaged with the
right amount, and the total must be right.

\paragraph{Build a playlist (7 standalone, 7 composed).} A friend has texted the rules for a playlist of the user's
starred songs; later texts change the rules. The agent must build the playlist in the music app and text the friend.
The playlist must contain exactly the songs the final rules allow.

\section{Agent harness}\label{app:prompt}
All models receive the same system prompt, tool schema and history format. At each step the model sees the request,
its own earlier replies and up to 20 recent screenshots (when the limit is reached, the oldest ten are replaced by a
short text note), and returns one call to a single phone tool with one of eleven actions: click, long press, swipe,
type, open an app, press a system button (Back, Home, Menu, Enter), send a key event, wait, answer, ask the user, or
terminate with a success or failure status. Coordinates are given on a 0--999 scale relative to the screenshot and
mapped to device pixels by the harness. A thin per-model adapter handles each model's coordinate
convention and tool-call syntax.
Thinking budgets and sampling are as in Section~\ref{sec:results}, with top-$k$ 20. An episode in which the agent acted on an incomplete tool call counts as a failure for every
model.

\onecolumn
\section*{Author contributions}
{\setlength{\tabcolsep}{0pt}\renewcommand{\arraystretch}{1.45}
\begin{tabular}{@{}>{\itshape}p{0.34\linewidth}p{0.66\linewidth}@{}}
Benchmark curation and training & Owais Aijaz \\
Task generation & Ahmer Tabassum, Nikhil Ranjan, Owais Aijaz \\
Training code & Aryan Seth, Siddhant Kharbanda \\
Evaluation infrastructure & Abhyuday Prasad, Amish Gupta \\
Project lead & Nikhil Ranjan \\
Supervision & Atmadeep Banerjee, Rahul Chhabra \\
\end{tabular}}

\section*{Citation}
\begin{tcolorbox}[enhanced,colback=white,colframe=hair,boxrule=0.4pt,sharp corners,left=10pt,right=10pt,top=6pt,bottom=6pt]
\ttfamily\footnotesize
@techreport\{sabi2026phonebench,\\
\hspace*{2em}title\ \ \ \ \ \ \ = \{\{PhoneBench\}: Phone Agents Ace Existing Benchmarks but Fail Everyday Tasks\},\\
\hspace*{2em}author\ \ \ \ \ \ = \{Aijaz, Owais and Ranjan, Nikhil and Tabassum, Ahmer and Gupta, Amish and\\
\hspace*{9.3em}Prasad, Abhyuday and Kharbanda, Siddhant and Seth, Aryan and\\
\hspace*{9.3em}Banerjee, Atmadeep and Chhabra, Rahul\},\\
\hspace*{2em}institution = \{Sabi AI Lab\},\\
\hspace*{2em}year\ \ \ \ \ \ \ \ = \{2026\},\\
\hspace*{2em}month\ \ \ \ \ \ \ = nov\\
\}
\end{tcolorbox}

\end{document}